\documentclass[fleqn,greek,10pt]{wlscirep}
\usepackage[utf8]{inputenc}
\usepackage[T1]{fontenc}
\usepackage{multirow}
\usepackage{longtable}
\title{Maistros: A Greek Large Language Model Adapted Through Knowledge Distillation From Large Reasoning Models}

\author[1,*]{Nikolaos Giarelis}
\author[1]{Charalampos Mastrokostas}
\author[1]{Nikos Karacapilidis}
\affil[1]{Industrial Management and Information Systems Lab, MEAD University of Patras, Rio Patras, Greece}

\affil[*]{giarelis@ceid.upatras.gr}

\begin{abstract}
Large Language Models (LLMs) have substantially advanced the field of Natural Language Processing (NLP), achieving state-of-the-art performance across a wide range of tasks. These improvements have been attributed, in part, to their emerging reasoning capabilities, which are enabled by large-scale training and increased model capacity. However, existing LLMs can generate erroneous responses when addressing complex queries that fall outside their training distribution, due to limited internal knowledge or the need for multi-step reasoning. To address these limitations, recent work has introduced large reasoning models (LRMs), which incorporate explicit internal reasoning processes to improve response accuracy. Additionally, state-of-the-art LRMs often comprise hundreds of billions of parameters and require several seconds per inference, even on advanced multi-GPU systems. These characteristics limit their practicality for deployment in conventional computing environments. Meanwhile, NLP research on multilingual LLMs continues to prioritize high-resource languages. However, these models exhibit limited performance in under-resourced languages, primarily due to insufficient language- and culture-specific training data. In this paper, we focus on Modern Greek, for which only a limited number of question answering (QA) datasets have been proposed, most of which are intended for model evaluation. To address this research gap in Greek QA, we make the following contributions: (i) We introduce \textit{CulturaQA}, a high-quality LRM-generated and human-curated dataset, for Greek LLM training and evaluation; (ii) a memory-efficient LLM evaluation framework adaptable to diverse languages and QA tasks; (iii) \textit{Maistros 8B}, a state-of-the-art open-weights Greek LLM developed via knowledge distillation and fine-tuning on \textit{CulturaQA}; and (iv) a comprehensive evaluation of nine LLMs across nine human-curated Greek QA datasets. We release our code, model, and data to support reproducibility.\end{abstract}
\begin{document}

\flushbottom
\maketitle
%
%
\thispagestyle{empty}

\section*{Introduction}

Large Language Models (LLMs) have significantly advanced the fields of Natural Language Processing (NLP), Artificial Intelligence (AI), and Deep Learning, achieving state-of-the-art performance across a wide range of natural language understanding and reasoning tasks \cite{minaee2025largelanguagemodelssurvey, naveed_2025}. LLMs, also referred to as Foundation Models, are trained on large-scale corpora using substantial computational resources (e.g., GPU clusters) and can subsequently be adapted to downstream tasks with comparatively limited resources \cite{bommasani2022opportunitiesrisksfoundationmodels}. Earlier LLMs such as \textit{GPT-3} \cite{brown_2020} and \textit{Llama-2} \cite{touvron2023llama2openfoundation} were primarily trained on English-centric corpora and exhibited limited multilingual capabilities. In contrast, recent models such as \textit{GPT-5} \cite{singh2025openaigpt5card} and \textit{Gemini 3} \cite{geminiteam2025geminifamilyhighlycapable} are trained on multilingual data and demonstrate improved cross-lingual and reasoning capabilities. A recent line of work further categorizes such models as Large Reasoning Models (LRMs), which generate extended intermediate reasoning traces to improve performance on complex tasks and reduce factual errors \cite{XU2025101370}. However, these capabilities come at an increased computational cost, including substantially larger model sizes, higher inference latency due to longer generated reasoning sequences, and a reliance on proprietary deployment infrastructures.

Despite these advances, multilingual LLMs continue to exhibit performance disparities between high-resource and under-resourced languages, largely due to imbalanced training data and limited coverage of linguistic and cultural variation \cite{shani2026rootsperformancedisparitymultilingual}. This issue is particularly evident in tasks requiring cultural or domain-specific knowledge, where models may produce incomplete or inaccurate outputs \cite{QIN2025101118}. In this context, Modern Greek remains an under-resourced language despite its linguistic complexity, including a distinct alphabet, rich morphology, and syntactic variability. Consequently, developing robust NLP systems for Greek remains challenging. Existing surveys highlight the scarcity of datasets, models, and systematic evaluations for Greek question answering (QA) \cite{BAKAGIANNI2025101313, papantoniou2024nlpgreeklanguagelonger, giarelis_2024_review}. While recent studies have introduced Greek QA datasets and benchmarking efforts \cite{roussis-etal-2025-krikri, zhang2026greekmmlunativesourcedmultitaskbenchmark, mastrokostas2026evaluatingmonolingualmultilinguallarge}, most datasets are primarily designed for evaluation rather than model training \cite{mastrokostas2026evaluatingmonolingualmultilinguallarge}.

Empirical findings further indicate that open-weights multilingual LLMs supporting Greek generally underperform compared to models specifically adapted for Greek, such as \textit{Krikri 8B} \cite{roussis-etal-2025-krikri}, while proprietary LLMs still achieve superior performance on Greek QA benchmarks \cite{zhang2026greekmmlunativesourcedmultitaskbenchmark, mastrokostas2026evaluatingmonolingualmultilinguallarge}. These observations highlight a persistent gap between open and proprietary models in under-resourced language settings. Motivated by these findings, this work investigates whether high-quality LRM-generated data can be leveraged to improve open-weights Greek language models. Specifically, using a human-in-the-loop process, we curate these data to mitigate linguistic and cultural inaccuracies. Finally, we employ supervised fine-tuning to distill knowledge into a compact model suitable for local deployment on commodity hardware. To address the identified gap in Greek QA, we make the following contributions:

\begin{itemize}
\item We introduce \textit{CulturaQA}, a synthetic and human-curated Greek QA dataset designed to support model training and evaluation.
\item We develop \textit{Maistros 8B}, a Greek-adapted open-weights LLM obtained via supervised fine-tuning of \textit{Ministral 3 8B} \cite{liu2026ministral3} on \textit{CulturaQA}.
\item We propose a memory-efficient and adaptable evaluation framework that supports  multiple-choice and open-ended QA tasks using the \textit{accuracy} and \textit{BERTScore} \cite{Zhang*2020BERTScore:} metrics, respectively.
\item We conduct a comprehensive evaluation of nine LLMs across nine human-curated Greek QA datasets.
\item We release the dataset, model, and code to support reproducibility (see the Data and Code Availability Statement sections).
\end{itemize}

\noindent This study investigates the following research questions (RQs):
\begin{itemize}

\item RQ1: Can human-curated LRM-generated data serve as a viable foundation for training and evaluating Greek QA systems?

\item RQ2: Does fine-tuning an open-weights LLM on \textit{CulturaQA} lead to measurable performance improvements on Greek QA benchmarks?

\item RQ3: How does the fine-tuned model compare to existing open-weights Greek and multilingual LLMs?

\item RQ4: To what extent can fine-tuned open-weights models approach the performance of proprietary LLMs on Greek QA tasks?

\end{itemize}

\section*{Related Work}

This section presents related works on language-adapted and general-purpose LLMs, resources for Greek QA, and common fine-tuning techniques. For this study, we consider LLMs with at least 7 billion parameters, as these models consistently outperform smaller ones in complex reasoning and language tasks\cite{minaee2025largelanguagemodelssurvey, naveed_2025}. We also employ their instruction-tuned model variants, as these are directly optimized for in-context learning and zero-shot NLP tasks. Throughout this paper, model sizes are denoted using standard abbreviations (e.g., 8B corresponds to 8 billion parameters).

\subsection*{General-purpose and Language-adapted LLM}

In this study, we consider several open-weights LLMs sorting them in two categories: (i) general-purpose and (ii) language-adapted. The first are typically made using large computing resources by for-profit AI organizations (e.g., Meta, Google, Mistral AI). The purpose of these LLMs is to excel in a variety of tasks, although prior works underscore their performance disparities across languages \cite{QIN2025101118, shani2026rootsperformancedisparitymultilingual}. On the other hand, language-adapted LLMs use the former ones as a basis to subsequently introduce language-, cultural- or domain-specific knowledge through LLM fine-tuning.

\textit{Llama 3.1 8B} \cite{grattafiori2024llama3herdmodels} and \textit{Gemma 3n E4B} \cite{gemmateam2025gemma3technicalreport} are two general-purpose LLMs that share a common training strategy. Both were pre-trained on large, multilingual corpora from the Internet, which also include mathematical and code reasoning data. Both of these models have learned many linguistic representations during pre-training and \textit{Gemma 3n E4B} officially supports a wide variety of languages. An important thing to note is that despite its confusing naming scheme, \textit{Gemma 3n’s} actual number of parameters is 8B.

\textit{Qwen 3 8B} \cite{yang2025qwen3technicalreport} is a general-purpose LLM that was pre-trained in a large and diverse corpus consisting of 119 languages and dialects. This corpus comprises high-quality content from diverse domains, including scientific books, code and reasoning tasks, multilingual texts, and synthetic data. The synthetic data were extracted from \textit{.pdf} documents using a \textit{Qwen} Vision Language model. \textit{Qwen-3 8B} is pre-trained on three different stages: (i) a general pre-training stage similar to previous models; (ii) a reasoning enhancing stage; and (iii) a training stage that improves its long context capabilities. Finally, Qwen-3 8B underwent post-training via knowledge distillation from a higher-capacity teacher model of the same family. This teacher model had been refined through reasoning-centric supervised fine-tuning and reinforcement learning.

\textit{Ministral 3 8B} \cite{liu2026ministral3} is a general-purpose LLM, that was trained using iterative layer pruning and knowledge distillation from a large pre-trained model of the same family. This model is later post-trained using instruction tuning and supervised fine-tuning from synthetic reasoning data. The authors' evaluation results indicate that the model performs at or above the level of counterparts with similar parameter counts, while requiring less training resources.

\textit{EuroLLM 9B v2} \cite{ramos2026eurollm22btechnicalreport} utilizes a specialized training strategy, where the model is trained with specific data per language, while incorporating math and code data for reasoning. At the same time, its authors have developed custom multilingual tokenizers to officially support 35 languages, including Greek. This model has several architectural and pre-training improvements over the first model version, and has incorporated synthetic math data from \textit{Gemma} and \textit{Llama} models. However, even in its revised version, the model still has an imbalanced number of language samples, where 18 European languages are critically underrepresented (less than 2\% of the training data).

\textit{Krikri 8B} \cite{roussis-etal-2025-krikri} is one of the first Greek-adapted LLMs, built on \textit{Llama 3 8B} \cite{grattafiori2024llama3herdmodels}. Krikri 8B was adapted from the base model, through extensive pre-training on Greek corpora, followed by Greek-specific instruction tuning, to improve its conversational capabilities. The authors' experimental results reported demonstrate its strong performance against other multilingual LLMs (including \textit{Llama 3 8B}) throughout many Greek NLP tasks.

\textit{Plutus 8B} \cite{peng-etal-2025-plutus} is a Greek financial LLM fine-tuned from \textit{Krikri 8B} on several financial tasks. It supports a variety of relevant tasks including QA, Named Entity Recognition, Text Summarization, Classification and Numerical Extraction. The authors report that it achieves state-of-the-art performance over several open-weights and proprietary models in the Greek Financial benchmark, which was introduced in the same work.

Some notable LLMs following similar training practices for other languages include: (i) \textit{LlaMandement-7B} \cite{gesnouin2024llamandementlargelanguagemodels}, a French adaptation of \textit{Llama 2 7B} that facilitates the Summarization of French Legislative Proposals; (ii) \textit{Llama-SEA-LION-8B-IT} and \textit{Gemma-SEA-LION-9B-IT} \cite{ong_2023} adapted from \textit{Llama 3 8B} and \textit{Gemma 2 9B} respectively, through additional pre-training, instruction tuning and supervised fine-tuning for South-East Asia languages; (iii) \textit{LlaMAntino-3-ANITA} \cite{polignano_advanced_2026}, an Italian adaptation of \textit{Llama 3 8B} using fine-tuning and reinforcement learning; (iv) \textit{GaMS3-12B-Instruct} \cite{vres2026buildingstronginstructionlanguage}, a Slovenian LLM adapted from \textit{Gemma 3 12B} using language-specific pre-training and fine-tuning. A common motif across these publications is that through additional linguistic or cultural knowledge distillation, the models perform much better than the original models used as a basis for several language-specific tasks.

\subsection*{Greek QA Datasets}
 
To ensure evaluation integrity, we focused exclusively on high-quality human-curated QA datasets for Greek. This approach avoids machine translation errors, which can negatively impact performance in NLP tasks \cite{graham_translationese_2019}. Using these criteria, we identified eight QA datasets suitable for the purposes of our study.

The \textit{Greek Medical MCQA} dataset \cite{voukoutis2024meltemiopenlargelanguage} comprises 2,034 medical QA pairs, where 1602 are reserved for training and the rest for validation. These QA pairs contain a question, five possible answer options with a single correct one. This dataset was extracted from the medical exams of the Hellenic National Academic Recognition and Information Center.

The \textit{Greek Truthful QA} \cite{voukoutis2024meltemiopenlargelanguage} contains 817 questions designed to measure whether LLMs can answer correctly when faced with misconceptions or false human beliefs. In contrast to other datasets, this one features a variable number of candidate answers for each QA pair. For our evaluation, we select its multiple-choice version, in the hardest difficulty setting (\textit{mc1\_targets}), where only a single answer is correct.

\textit{INCLUDE} \cite{romanou2025include} is a large-scale multilingual dataset spanning 44 languages. This dataset is collected from local academic and professional exams, and facilitates per-language LLM evaluation of regional and domain-specific knowledge. In our experiments, we utilize the Greek subset (552 QA pairs), where each pair consists of a question, four possible answers and a single correct one.

\textit{Greek ASEP MCQA} \cite{mcqa_greek_asep} comprises 1,200 multiple-choice QA pairs, sourced from the Greek Supreme Council for Civil Personnel Selection (ASEP) exams. This dataset spans various topics, including Greek history, law, politics, public administration and e-governance. This dataset has the same QA structure as \textit{INCLUDE}.

\textit{GPCR} \cite{gprc_greek}, also known as the Greek Physical Commonsense Reasoning dataset, includes manually-annotated 208 samples, similar to \textit{PIQA} \cite{chang2025globalpiqaevaluatingphysical}. These samples contain a question, two candidate answers exhibiting near-identical lexical composition, with a single one marked as correct. Approximately 40\% of the samples are regionally or culturally specific, and cannot be easily rendered into English.

\textit{Plutus QA} \cite{peng-etal-2025-plutus} consists of 540 Greek financial QA pairs. This dataset was derived from real-world financial documents (i.e., annual reports, article headlines and exam questions) and was annotated by several financial and linguistics experts. \textit{Plutus QA} has a train, validation and test splits that comprise 267, 48 and 225 pairs respectively. Each QA pair has a question, additional context and a set of multiple-choice answers, with a single correct one.

\textit{Demos QA} \cite{mastrokostas2026evaluatingmonolingualmultilinguallarge} is a Greek QA dataset, comprising 600 questions and community-reviewed answers from Greek social media. Each QA pair has a question and four candidate answers, as well as a selected best answer \textit{DemosQA’s} answers are ranked based on community voting, with the highest-upvoted response designated as the reference answer.

\textit{Greek MMLU} \cite{zhang2026greekmmlunativesourcedmultitaskbenchmark} is a QA dataset that assesses massive multitask Greek understanding. It encompasses 45 subjects with 21,805 original QA pairs curated from real-world educational and professional assessments. The present study focuses on its Greek-specific subset comprising 3,660 question-answer (QA) pairs. These pairs necessitate a higher level of Hellenic cultural and linguistic knowledge.

Collectively, these eight human-curated datasets offer a valuable foundation to evaluate LLMs supporting Greek in varying domains. However, most of them lack training samples across a multitude of topics related to Greece and its culture; therefore, this motivates us to create our dataset \textit{CulturaQA}, which is elaborated in a following section.

\subsection*{Supervised Fine-Tuning}
LLMs comprise many dense weight matrices, which are used to infer the next tokens in the sequence, given an input. In order to instill new knowledge in the model, it is required to update these weights. The typical learning paradigm is full fine-tuning. Let us consider a single dense matrix $W_0 \in \mathbb{R}^{d \times k}$ from the LLM. To update this matrix, we would have to produce a new one $\Delta$W: 

\begin{equation}\label{eq:1}
    W = W_0 + \Delta W
\end{equation}

where $\Delta$W would be the learned weight updates, which are produced by recalculating the gradients for the entire parameter space \textit{(d x k)}. Several studies have shown this to be computationally expensive, given the large number of parameters of LLMs, where multiple GPUs with extremely large memory are required \cite{hu2022lora, mao_survey_2024, biderman2024lora}. Thus, to mitigate this issue, Hu et al. \cite{hu2022lora} introduce a different fine-tuning paradigm; Low-Rank Adaptation (LoRA). Instead of $\Delta$W, LoRA calculates two smaller low-rank matrices $A \in \mathbb{R}^{d \times r}$ and $B \in \mathbb{R}^{r \times k}$ that alter equation (1) as such: 

\begin{equation}\label{eq:2}
    W = W_0 + sAB
\end{equation}

where $r \ll \min\{d, k\}$ and s is the scaling factor which controls the impact of weight updates: 

\begin{equation} \label{eq:3}
    s = \frac{\alpha}{r}
\end{equation}

The scaling factor is typically set to 2.0, to improve training stability and robustness \cite{biderman2024lora, shuttleworth2025lora}. This is achieved by setting $\alpha$ to be twice as large as the selected rank parameter. The matrices A and B have their elements initially set to a random Gaussian distribution and 0 respectively. LoRA optimizes efficiency by learning low-rank matrices, which enable fine-tuning while only updating a fraction of the original LLM's parameters. Despite, this smaller update, LoRA has been shown to be nearly equivalent with full fine-tuning, while mitigating the effects of catastrophic forgetting of previous model knowledge \cite{biderman2024lora, shuttleworth2025lora}. Considering that an LLM has multiple layers (L) and trainable modules (M), we can define equation \ref{eq:2} as such, where each layer $l \in L$ and $m \in M$: 

\begin{equation} \label{eq:4}
    W_{(l,m)} = W_{0(l,m)} + \frac{\alpha}{r} (B_{(l,m)}A_{(l,m)})
\end{equation}

Both in the case of full and LoRA fine-tuning an LLM learns by minimizing the following loss function \cite{aakanksha_2023}: 

\begin{equation} \label{eq:5}
    L = - \frac{1}{T} \sum_{t=1}^{T} \log P(x_t \mid x_{<t})
\end{equation}

where T is the total number of tokens, while $x_t$ and $x_{<t}$ are the current and preceding tokens of the input sequence X. Essentially, the model learns to minimize the loss (or maximize the probability) of correctly predicting the next token in the sequence from previous ones, based on the training examples. However, Touvron et al. \cite{touvron2023llama2openfoundation} suggested an improved version of this loss function: 

\begin{equation} \label{eq:6}
    L = - \frac{1}{\sum_{t=1}^{T} m_t} \sum_{t=1}^{T} m_t \log P(x_t \mid x_{<t})
\end{equation}

In this version, a token-level binary mask $m_t \in \{0, 1\}$ is introduced. This mask is assigned to 0 for every token that is part of the user or system prompt, (and 1 otherwise), thus these tokens are omitted from the loss calculation. By computing gradients only on the assistant's generated response tokens, the model is prevented from learning the user and system prompt distributions, thus significantly enhancing its chat and instruction-following capabilities. In this work, we are fine-tuning our proposed model based on equations \ref{eq:4} and \ref{eq:6}. For more information about our training see the Model Training and Validation Section. 

\section*{The CulturaQA Dataset}
In this section, we introduce \textit{CulturaQA}, a synthetic and human curated dataset that captures knowledge from Greek culture. \textit{CulturaQA} encompasses a plethora of topics across several domains including Greek art, history, mythology, politics, economy, tourism, food, health, science, sports, education and law, thus providing a valuable resource for training, validating and evaluating models on the nuances of Greek Culture, as well as advancing language understanding research within culturally grounded contexts. 

\begin{figure}[ht]
\centering
\includegraphics[height=0.37\textheight]{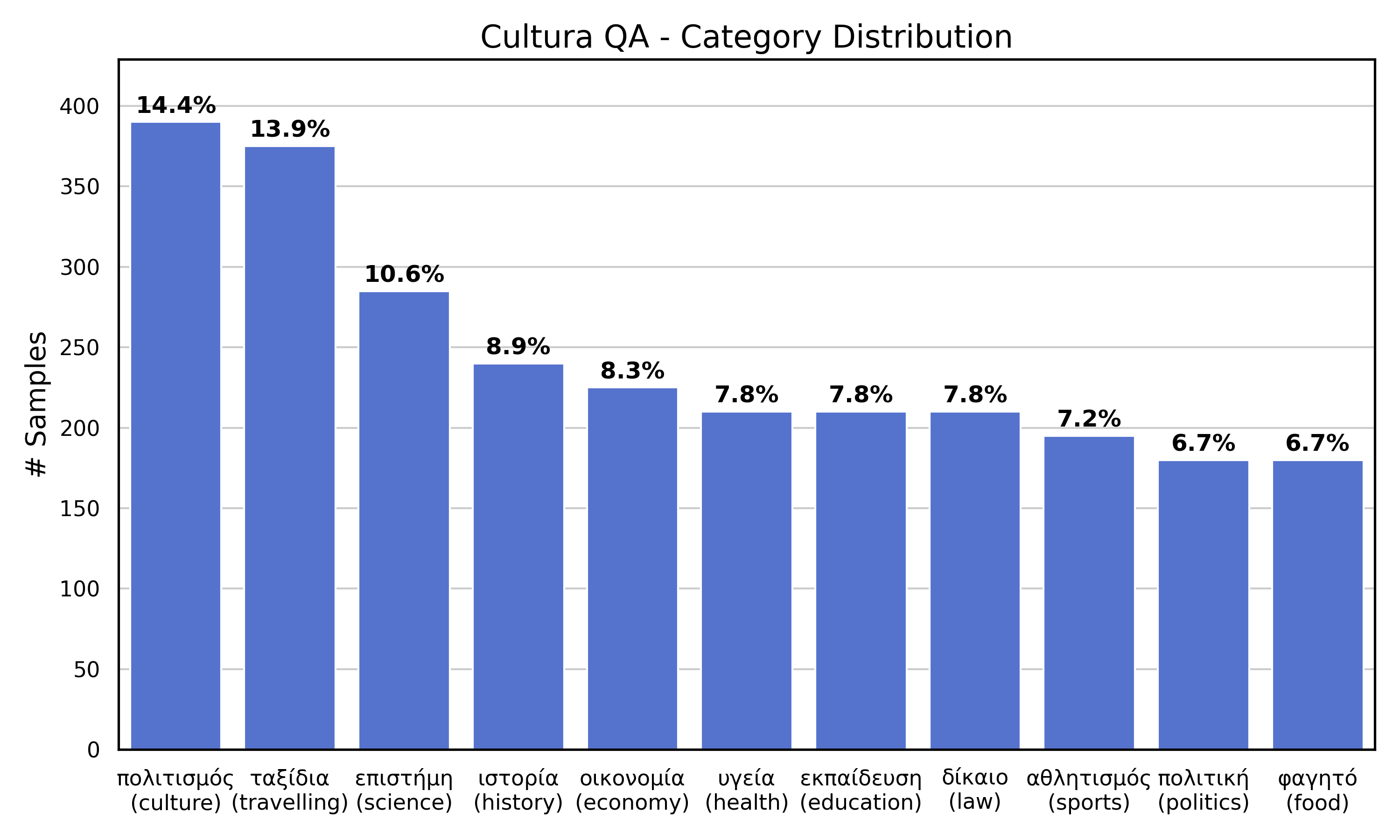}
\caption{CulturaQA's sample distribution per category. The y-axis measures the number of samples; the percentage of samples is written in each bar.}
\label{fig:1}
\end{figure}

To create \textit{CulturaQA}, we manually curated a list of 180 Greek keyphrases (for the exact phrases, see the code repository) that were grouped into eleven categories. These categories are “\textgreek{πολιτισμός}” (“civilization”), “\textgreek{ταξίδια}” (“travelling”), "\textgreek{πολιτική}" (“politics”), “\textgreek{οικονομία}” (“economy”), “\textgreek{επιστήμη}” (“science”), “\textgreek{υγεία}” (“health”), “\textgreek{αθλητισμός}” (“sports”), “\textgreek{εκπαίδευση}” (“education”), “\textgreek{ιστορία}” (“history”), “\textgreek{φαγητό}” (“food”) and “\textgreek{δίκαιο}” (“law”). For each keyphrase, we generated 15 questions using \textit{GPT-5}, which were answered one-by-one by the same model. Each QA pair is accompanied by a unique ID and the category that it belongs. For the exact dataset creation prompts, please refer to the Appendix. This procedure resulted in a dataset comprising 2,700 Greek question-answer pairs. For the sample distribution per category for CulturaQA (see Figure \ref{fig:1}).

We then manually pre-processed the dataset to ensure its linguistic quality, as well as remove any erroneous answers (hallucinations) and mitigate potential cultural and historical biases. During this step, we correct several linguistic errors, for example Greek words: (i) written in the wrong grammar inflection; (ii) containing english characters; (iii) that were misspelled or (iv) erroneously written in polytonic. We also corrected syntax and translation errors, where a few non-Greek words were generated, while an equivalent Greek term existed. We added the missing full forms for abbreviations that were used to refer to Greek organizations. We corrected several factual hallucinations and we removed suggestions, where the model was asking follow-up questions (e.g., “Should I produce a table, report, file, etc. to compile the above information?”). Finally, we removed the commonly occurring phrase, given a lack of reference for time or country, when in fact, there was a time reference (e.g., the last five years, or the current year) or a local reference (e.g., Greece) in the question. 

During post-processing, we utilized code to remove unnecessary whitespace and replace the English with the Greek question mark from each QA pair. We also divided the dataset into training, validation and testing splits using stratifying sampling based on the categories, to ensure that all splits have a similar distribution across categories. We also analyzed \textit{CulturaQA} and compared it with other datasets (see Table \ref{tab:dataset_stats}). 

\begin{table}[ht]
\centering
\resizebox{\linewidth}{!}{
\begin{tabular}{|l|c|c|l|c|c|c|c|c|c|c|}
\hline
\textbf{Dataset} & \textbf{\#Docs} & \textbf{Train, Val, Test splits} & \textbf{Type} & \textbf{P5} & \textbf{P25} & \textbf{P50} & \textbf{Mean} & \textbf{P75} & \textbf{P95} & \textbf{P99} \\
\hline
\multirow{2}{*}{CulturaQA} & \multirow{2}{*}{2700} & \multirow{2}{*}{2000, 200, 500} & Question & 11 & 14 & 17 & 17.81 & 20 & 26 & 32.1 \\
\cline{4-11}
 & & & Answer & 26 & 98.75 & 178 & 201.7 & 295 & 451.35 & 557.31 \\
\hline
Greek MMLU & \multirow{2}{*}{3660} & \multirow{2}{*}{0, 0, 3660} & Question & 6 & 8 & 12 & 56.08 & 18 & 308 & 375.41 \\
\cline{4-11}
(Greek Specific) & & & Answer & 1 & 1 & 2 & 3.03 & 4 & 10 & 15.41 \\
\hline
Greek Medical & \multirow{2}{*}{2032} & \multirow{2}{*}{1602, 432, 0} & Question & 3 & 6 & 9 & 9.96 & 11.25 & 21 & 27 \\
\cline{4-11}
MCQA & & & Answer & 1 & 2 & 3.5 & 4.67 & 6.0 & 12 & 17.7 \\
\hline
Greek ASEP & \multirow{2}{*}{1200} & \multirow{2}{*}{0, 0, 1200} & Question & 4 & 6 & 10 & 11.4 & 14 & 26 & 37.01 \\
\cline{4-11}
MCQA & & & Answer & 2 & 4 & 7 & 8.38 & 11.25 & 21 & 27 \\
\hline
Greek Truthful & \multirow{2}{*}{817} & \multirow{2}{*}{0, 0, 817} & Question & 5 & 7 & 9 & 11.03 & 13 & 22 & 40.68 \\
\cline{4-11}
QA & & & Answer & 2.8 & 7 & 10 & 10.02 & 13 & 18 & 21.84 \\
\hline
\multirow{2}{*}{DemosQA} & \multirow{2}{*}{600} & \multirow{2}{*}{0, 0, 600} & Question & 26 & 53 & 84.5 & 103.04 & 132.25 & 243 & 347.7 \\
\cline{4-11}
 & & & Answer & 11 & 31 & 54.5 & 80.22 & 105 & 222 & 362.04 \\
\hline
Include & \multirow{2}{*}{552} & \multirow{2}{*}{0, 0, 552} & Question & 5 & 9 & 13 & 22.8 & 27.25 & 75.9 & 129.96 \\
\cline{4-11}
(Greek) & & & Answer & 1 & 3 & 5 & 7.23 & 9 & 20 & 35 \\
\hline
\multirow{2}{*}{GPCR} & \multirow{2}{*}{208} & \multirow{2}{*}{0, 0, 208} & Question & 5 & 7 & 10 & 11.76 & 13 & 25 & 45.58 \\
\cline{4-11}
 & & & Answer & 2 & 4 & 7 & 10.64 & 12 & 32.30 & 47.65 \\
\hline
\multirow{2}{*}{Plutus QA} & \multirow{2}{*}{540} & \multirow{2}{*}{267, 48, 225} & Question & 3.2 & 7 & 11 & 13.45 & 16 & 34 & 54 \\
\cline{4-11}
 & & & Answer & 1 & 3 & 7 & 11.01 & 12 & 36.8 & 69.84 \\
\hline
\end{tabular}
}
\caption{\label{tab:dataset_stats}Number of documents and splits per dataset, with a statistical overview of their question and answer word counts.}
\end{table}

\section*{LLM Post-training, Validation and Evaluation}
\begin{figure}[ht]
\centering
\includegraphics[height=0.35\textheight]{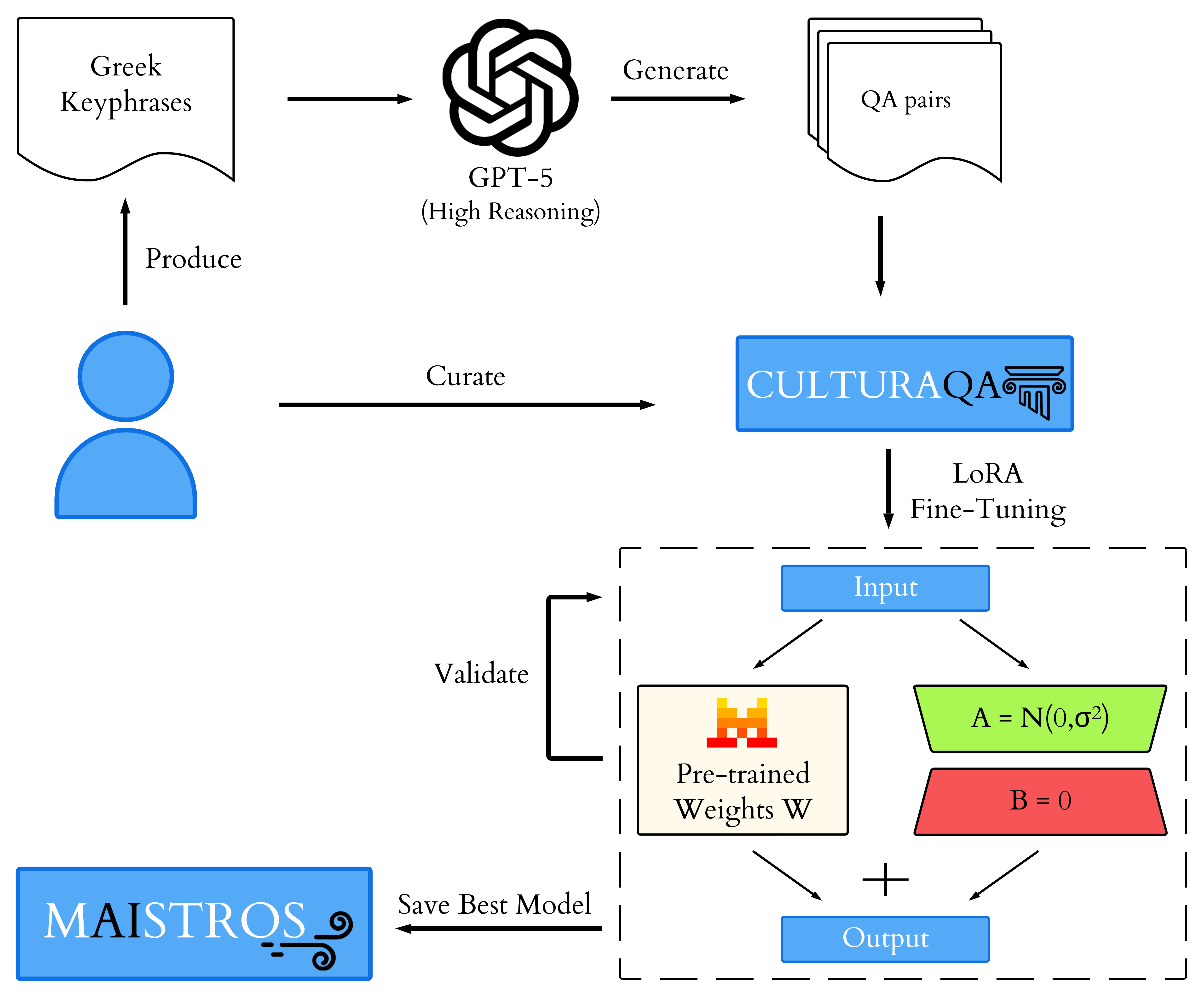}
\caption{The overall approach for training \textit{Maistros 8B}.}
\label{fig:2}
\end{figure}

This section presents our technical setup, the post-training and validation of the proposed model \textit{Maistros 8B}, as well as our empirical evaluation. The general approach is illustrated in Figure \ref{fig:2}.

\begin{table}[t]
\centering
\begin{tabular}{|l|c|c|}
\hline
\textbf{LLM} & \textbf{Full Model Name} & \textbf{Open weights} \\
\hline
GPT-5 mini & openai/gpt-5-mini-2025-08-07 & - \\
\hline
Gemini 3 Flash & google/gemini-3-flash-preview & - \\
\hline
Ministral 3 8B & mistralai/Ministral-3-8B-Instruct-2512-BF16 & \checkmark \\
\hline
Krikri 8B & ilsp/Llama-Krikri-8B-Instruct & \checkmark \\
\hline
Plutus 8B & TheFinAI/plutus-8B-instruct & \checkmark \\
\hline
EuroLLM 9B v2 & utter-project/EuroLLM-9B-Instruct-2512 & \checkmark \\
\hline
Gemma 3n E4B & google/gemma-3n-E4B-it & \checkmark \\
\hline
Qwen 3 8B & Qwen/Qwen3-8B & \checkmark \\
\hline
Maistros 8B & IMISLab/Maistros 8B-Instruct & \checkmark \\
\hline
\end{tabular}
\caption{\label{tab:2}Considered LLMs for the experiments.}
\end{table}

\subsection*{Technical Setup}

In terms of hardware specifications, for LLM fine-tuning, we utilized a cloud server with 16 logical cores CPU, 128 GB of RAM and an \textit{Nvidia L40S} 48 GB VRAM GPU, whereas for LLM evaluation we utilized a local server with an \textit{Intel Core i9-12900K} 20 logical cores CPU, 64 GB of RAM and an \textit{Nvidia RTX A4000} 16 GB VRAM GPU. 

Regarding software specifications, we utilized several HuggingFace \cite{wolf-etal-2020-transformers} libraries (i.e., \textit{Transformers}, \textit{TRL}, \textit{PEFT}) for LoRA Fine-tuning, as well as \textit{Bitsandbytes} to load models with 4-bit quantization as to reduce memory requirements for our training and the evaluation of open-weights LLMs. Morever, we utilized the GenAI APIs from OpenAI and Google, to access \textit{GPT-5}, for the synthetic data creation step, as well as \textit{GPT-5 mini} and \textit{Gemini 3 Flash} for model evaluation. Finally, we use \textit{scikit-learn’s} accuracy metric for multiple-choice QA and \textit{BERTScore} for open-ended QA. For reproducibility purposes, we set a fixed random seed (42) and for model evaluation we also utilize greedy decoding, which is equivalent to model temperature 0.0 \cite{renze-2024-effect}.

To preserve a controlled experimental setup, we selected open-weights LLMs in the same computational class (ranging from 7B to 10B parameters), while including the smallest available versions of state-of-the-art proprietary LLMs. We also utilize the most recent model versions (e.g., \textit{Gemma 3}). We exclude models that were shown to have bad performance in Greek QA tasks from previous empirical studies \cite{mastrokostas2026evaluatingmonolingualmultilinguallarge, zhang2026greekmmlunativesourcedmultitaskbenchmark}. Overall, the models evaluated in this study are compiled in Table \ref{tab:2}.

\subsection*{Model Training and Validation}

To develop \textit{Maistros 8B}, we post-trained \textit{Ministral-3-8B-Instruct} on \textit{CulturaQA} using LoRA fine-tuning. For our training setup, we utilize a mixed precision environment, where the model is loaded initially in full precision (\textit{BFloat16}) and then it is \textit{4-bit} quantized to a normalized float (\textit{NF4}) with double quantization to reduce the memory requirements even further \cite{dettmers_2023_qlora}. Following best practices from Touvron et al. \cite{touvron2023llama2openfoundation}, we also calculate the cross-entropy training loss for the answer, while question and system prompts are masked and thus excluded from the loss calculation. The aim of this practice is to update the model based on the answer and not learned fixed patterns that occur at the question and instructions of the system prompt. This has the added benefit that the model does not learn to predict the question. Thus after fine-tuning, it directly answers the question instead of generating it again. We experimented with various training hyperparameters (see Table \ref{tab:3}) and we selected the best ones based on the training and validation losses. The ones that are elaborated below are the best ones for the final training run.

We train the model for 4 epochs using a learning rate of 2e-5 and a cosine learning rate scheduler, with a batch size of 2 and gradient accumulation steps set to 8. This gives us a larger effective batch size of 16, which improves training stability \cite{smith_2018}. To further improve stability, we utilized global gradient clipping with a maximum L2 norm threshold of 1.0 \cite{Zhang2020Why}). Overall, our goal is to keep the batch size small, due to the low number of training samples. To further reduce memory usage, we employed the \textit{8-bit AdamW} optimizer \cite{dettmers20228bitoptimizersblockwisequantization} with its default hyperparameters ($\beta_1 = 0.9$, $\beta_2 = 0.999$, and $\epsilon = 1e{-8}$). This optimizer reduces the memory usage by 75\% by utilizing block-wise dynamic quantization (from \textit{32-bit} to \textit{8-bit}).

We add LoRA adapters across all attention and feed forward layers of the model. For LoRA, we set the rank to 16 and the LoRA alpha ($\alpha$) to 32; this gives us a scaling factor of 2.0, which has been shown to improve training stability and robustness \cite{biderman2024lora, shuttleworth2025lora}. To avoid overfitting, we set both the dropout for every LoRA layer and weight decay to 0.1. For our training, we set the max sequence length to 3269. This is calculated by counting the number of tokens of the longest training sample; we also pad sequences to the length of the longest one in each training batch to reduce computational overhead. The total number of training steps is 500 calculated using the following equation and the warm-up steps are set to 62 (8\% of total):
\begin{equation} \label{eq:training_steps}
    \left\lfloor \frac{2000 \text{ training samples}}{16 \text{ (effective batch size)}} \times 4 \text{ epochs} \right\rfloor = 500 \text{ steps}
\end{equation}

\begin{table}[ht]
\centering
\begin{tabular}{|l|c|}
\hline
\textbf{Training Hyperparameter} & \textbf{Values} \\
\hline
Train Batch size & \{1, \textbf{2}\} \\
\hline
Gradient Accumulation Steps & \{4, \textbf{8}\} \\
\hline
Learning Rate & \{1e-4, 9e-5, 5e-5, 3e-5, \textbf{2e-5}\} \\
\hline
Weight decay & \{0, 0.01, \textbf{0.1}\} \\
\hline
LoRA Rank & \{8, 16, \textbf{32}\} \\
\hline
LoRA Dropout & \{0, 0.05, \textbf{0.1}\} \\
\hline
Warmup steps & \{25, 50, \textbf{62}\} \\
\hline
Learning Rate Scheduler & \{linear, \textbf{cosine}\} \\
\hline
Target modules & \begin{tabular}{@{}c@{}}
Attention mechanism: Query (\textbf{q\_proj}), Key (\textbf{k\_proj}), Value(\textbf{v\_proj}) \\
The Output projection of the attention layer: (\textbf{o\_proj}) \\
Feed-Forward Network blocks: (\textbf{gate\_proj}, \textbf{up\_proj}, \textbf{down\_proj})
\end{tabular} \\
\hline
\end{tabular}
\caption{\label{tab:3}Training hyperparameters tested for fine-tuning; the best parameters are marked in bold.}
\end{table}

\begin{figure}[ht]
\centering
\includegraphics[height=0.37\textheight]{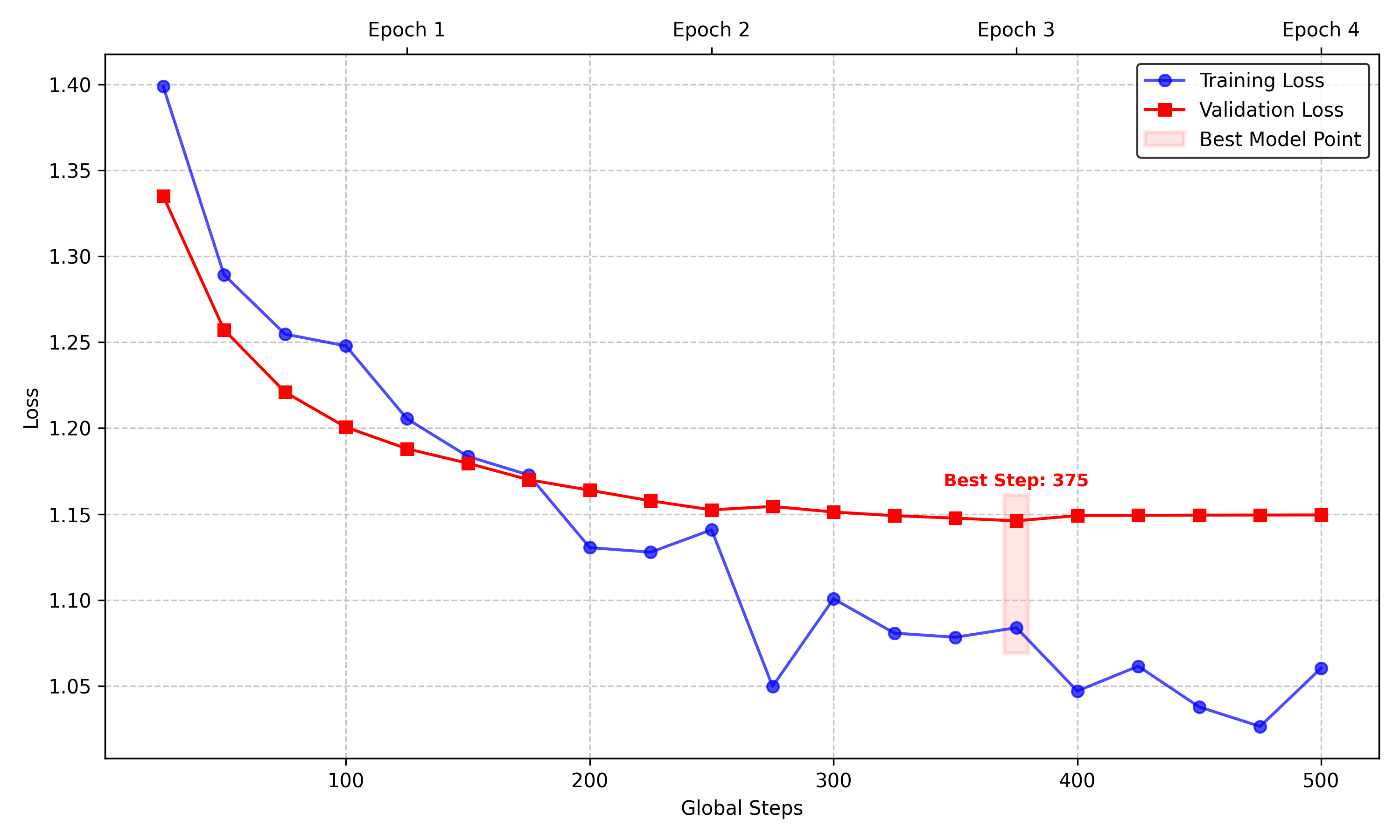}
\caption{Training and Evaluation loss over steps}
\label{fig:3}
\end{figure}

The amount of total steps is equally divided among epochs, thus each epoch comprises 125 steps. To select the best model checkpoint for the final training run, we calculated the training and validation losses and the best LoRA adapters for the model are trained on step 375 (epoch 3) where the lowest validation and training losses are achieved. This is visualized in Figure \ref{fig:3}. Finally, we merge the LoRA adapter weights from epoch 3 to the weights of the base model to produce \textit{Maistros 8B}. 

\subsection*{Experiments}
This section elaborates on our evaluation framework and the empirical results. To obtain these results, we carried out a series of experiments that measure the performance of the considered LLMs for Greek QA tasks. The point of these tasks is to reveal model effectiveness in understanding and generating accurate responses to Greek questions across diverse topics. We evaluated the considered LLMs (see Table \ref{tab:2}) on eight multiple-choice and one open-ended QA task. For multiple-choice tasks, we extract the correct answer from the LLM response using rule-based parsing and regular expressions, since instruction-tuned models often include explanatory text alongside their selected answer. If a valid answer could not be extracted, it was labeled as “\textit{No match}”. For the open-ended QA task (Cultura QA), the models generate a full answer, which is then evaluated using the \textit{BERTScore F1} metric (\%) against the reference answer. For the exact instruction prompts used in the evaluation, see Appendix A. One final note is that our framework standardizes different dataset formats into the same format as to enable the comparative performance evaluation. 

\begin{table}[ht]
\centering
\resizebox{\linewidth}{!}{
\begin{tabular}{|l|c|c|c|c|c|c|c|c|c|}
\hline
\textbf{Score \%} & \textbf{DemosQA} & \textbf{GPCR} & \textbf{INCLUDE} & \begin{tabular}{@{}c@{}}\textbf{Greek ASEP} \\ \textbf{MCQA}\end{tabular}& \begin{tabular}{@{}c@{}}\textbf{Greek Medical} \\ \textbf{MCQA}\end{tabular} & \textbf{Plutus QA} & \begin{tabular}{@{}c@{}}\textbf{Greek} \\ \textbf{Truthful QA}\end{tabular} & \begin{tabular}{@{}c@{}}\textbf{Greek MMLU} \\ \textbf{(Greek-specific)}\end{tabular} & \textbf{CulturaQA}  \\
\hline
\multicolumn{10}{|c|}{\textbf{Proprietary Models}} \\
\hline
Gemini 3 flash & \textbf{55.67} & \textbf{88.46} & \textbf{88.77} & \textbf{94.75} & \textbf{92.82} & \textbf{89.78} & \textbf{88.62} & \textbf{95.03} & 73.97 \\
\hline
GPT-5 mini & 53.00 & 77.40 & 74.46 & 78.92 & 78.01 & 76.89 & 75.89 & 87.49 & \textbf{75.09} \\
\hline 
\multicolumn{10}{|c|}{\textbf{Open-Weights Models}}\\
\hline
Maistros 8B (Ours) & 50.83 & \textbf{64.42} & \textbf{58.70} & \textbf{67.25} & \textbf{49.54} & \textbf{73.33} & 53.37 & \textbf{78.17} & \textbf{71.99}\\
\hline
Ministral 3 8B & \textbf{51.67} & 59.62 & 54.17 & 63.25 & 47.92 & 65.33 & 52.51 & 76.23 & 71.03 \\
\hline
Krikri 8B & 49.50 & 54.81 & 50.54 & 63.08 & 45.37 & 64.44 & \textbf{54.83} & 71.04 & 71.31 \\
\hline
Plutus 8B & 45.67 & 50.00 & 48.37 & 62.92 & 39.35 & 57.33 & 34.52 & 70.38 & 67.44\\
\hline
EuroLLM v2 9B & 41.50 & 53.85 & 39.13 & 46.08 & 31.71 & 42.67 & 36.72 & 58.17 & 70.33\\
\hline
Gemma 3n E4B & 47.17 & 60.10 & 50.00 & 57.75 & 43.75 & 53.78 & 46.76 & 71.39 & 69.10\\
\hline
Qwen 3 8B & 48.83 & 31.73 & 49.28 & 54.58 & 36.64 & 63.56 & 42.72 & 67.57 & 68.73\\
\hline 

\end{tabular}
}
\caption{\label{tab:4}Empirical Greek QA results. We report macro accuracy for multiple-choice QA datasets, and macro BERTScore F1 for CulturaQA. The best results for proprietary and open-weights models are highlighted in bold.}
\end{table}

As shown in Table \ref{tab:4}, for most QA datasets, the best performance is achieved by the proprietary models, with \textit{Gemini 3 Flash} outperforming \textit{GPT-5-mini} more than 10\% percent across most of them. A notable exception is \textit{CulturaQA}, which was generated using GPT-5 (the parent model of \textit{GPT-5 mini}); thus, in this dataset the former outperforms \textit{Gemini 3 Flash} by a small margin.
Regarding open-weights models, \textit{Maistros 8B} achieves the state-of-the-art scores in most datasets with meaningful improvements over the base model. Two notable exceptions are \textit{Demos QA} and \textit{Greek Truthful}, where \textit{Maistros 8B} is outperformed by \textit{Ministral 3 8B} and \textit{Krikri 8B} respectively, albeit by a small margin (less than 2\%). One of the most interesting results, was model performance on \textit{Plutus QA}, a domain-specific dataset (Greek Economy), where \textit{Maistros 8B} outperformed all open-weights models, while simultaneously attaining a similar accuracy score with \textit{GPT-5 mini} (-3.56\% difference). Similarly, for the test set of \textit{CulturaQA}, \textit{Maistros 8B} outperformed all open-weights models, while simultaneously attaining a similar score with the proprietary models.

To assess the statistical significance of the observed score improvements, we compared \textit{Maistros 8B} against the base model across the above datasets. For the multiple-choice QA datasets, we encoded outputs as paired binary data, where if a model selected the reference answer, we labelled this as 1 and 0, otherwise. Statistical significance was measured using the exact binomial \textit{McNemar’s test}. To quantify uncertainty in the estimated effect sizes, we employed bootstrap resampling to derive 95\% \textit{confidence intervals (CI)} for the accuracy differences. Specifically, 10,000 bootstrap samples were drawn with replacement from the paired binary data. For the Cultura QA dataset, we utilized the Wilcoxon signed-rank test, given the continuous nature of the BERTScore F1 scores, and similarly 10,000 bootstrap samples were drawn to derive the corresponding 95\% \textit{CI}.

In both cases, we consider the score improvement as statistically significant only if the \textit{p-value} <= 0.05 and the resulting 95\% \textit{CI} was strictly positive. The reason we run the above tests is that they are recommended to measure the statistical significance of metrics such as accuracy and F1 scores for small NLP datasets \cite{dror-etal-2018-hitchhikers}. The results of these tests are collected in the following tables. Specifically, in Table \ref{tab:5}, we observe the exact accuracy improvements of \textit{Maistros 8B} over the base model, where the former achieves a statistically significant improvement in 5 out of 9 datasets. In Tables \ref{tab:6} and \ref{tab:7}, we report the \textit{p-value} and 95\% \textit{CI} for \textit{Maistros 8B} compared against other open-weights models in every dataset. In these tables, we observe that the proposed model achieves statistical significant improvements across most datasets. Some notable exceptions include the \textit{Demos QA} dataset, where \textit{Maistros 8B} does not achieve a statistical significant improvement against \textit{Krikri 8B}, \textit{Gemma 3n E4B} and \textit{Qwen 3 8B}. This is also the case for the \textit{Greek Medical MCQA} and \textit{Greek Truthful QA} datasets, where the proposed model does not improve statistically against \textit{Krikri 8B}.

\begin{table}[ht]
\centering
\begin{tabular}{|l|c|c|c|}
\hline
\textbf{Dataset} & \textbf{Score Improvement} & \textbf{P-Value} & \textbf{95\% CI} \\
\hline
DemosQA & -0.83\% & 0.635 & [-3.67\%, 2.00\%] \\
\hline
Greek PCR & 4.81\% & 0.064 & [0.0\%, 9.62\%] \\
\hline
INCLUDE & \textbf{4.53\%} & \textbf{0.014} & \textbf{[1.09\%, 7.97\%]} \\
\hline
Greek ASEP MCQA & \textbf{4.00\%} & \textbf{0.001} & \textbf{[1.75\%, 6.33\%]} \\
\hline
Greek Medical MCQA & 1.62\% & 0.470 & [-2.08\%, 5.32\%] \\
\hline
Plutus QA & \textbf{8.00\%} & \textbf{0.001} & \textbf{[3.56\%, 12.44\%]} \\
\hline
Greek Truthful QA & 0.86\% & 0.562 & [-1.59\%, 3.30\%] \\
\hline
Greek MMLU (Greek-Specific) & \textbf{1.94\%} & \textbf{0.000} & \textbf{[1.01\%, 2.87\%]} \\
\hline
CulturaQA & \textbf{0.97\%} & \textbf{0.000} & \textbf{[0.49\%, 1.46\%]} \\
\hline
\end{tabular}
\caption{\label{tab:5}Score improvements of Maistros 8B over the base model and results from the statistical significance tests.}
\end{table}

\begin{table}[ht]
\centering
\resizebox{\linewidth}{!}{
\begin{tabular}{|l|c|c|c|c|c|c|c|c|c|}
\hline
\textbf{p-value} & \textbf{DemosQA} & \textbf{GPCR} & \textbf{INCLUDE} & \begin{tabular}{@{}c@{}}\textbf{Greek ASEP} \\ \textbf{MCQA}\end{tabular}& \begin{tabular}{@{}c@{}}\textbf{Greek Medical} \\ \textbf{MCQA}\end{tabular} & \textbf{Plutus QA} & \begin{tabular}{@{}c@{}}\textbf{Greek} \\ \textbf{Truthful QA}\end{tabular} & \begin{tabular}{@{}c@{}}\textbf{Greek MMLU} \\ \textbf{(Greek-specific)}\end{tabular} & \textbf{CulturaQA}  \\
\hline
Krikri 8B & 0.575 & \textbf{0.010} & \textbf{0.001} & \textbf{0.006} & 0.153 & \textbf{0.017} & 0.495 & \textbf{0.000} & \textbf{0.002} \\
\hline
Plutus 8B & \textbf{0.030} & \textbf{0.000} & \textbf{0.000} & \textbf{0.004} & \textbf{0.000} & \textbf{0.000} & \textbf{0.000} & \textbf{0.000} & \textbf{0.000}\\
\hline
EuroLLM v2 9B & \textbf{0.001} & \textbf{0.033} & \textbf{0.000} & \textbf{0.000} & \textbf{0.000} & \textbf{0.000} & \textbf{0.000} & \textbf{0.000} & \textbf{0.000}\\
\hline
Gemma 3n E4B & 0.117 & 0.362 & \textbf{0.000} & \textbf{0.000} & \textbf{0.038} & \textbf{0.000} & \textbf{0.000} & \textbf{0.000} & \textbf{0.000}\\
\hline
Qwen 3 8B & 0.399 & \textbf{0.000} & \textbf{0.000} & \textbf{0.000} & \textbf{0.000} & \textbf{0.006} & \textbf{0.000} & \textbf{0.000} & \textbf{0.000}\\
\hline 

\end{tabular}
}
\caption{\label{tab:6}Statistical significancy tests for Maistros 8B against the other open-weights LLMs. Results with p-value < 0.05 are highlighted in bold.}
\end{table}

\begin{table}[ht]
\centering
\resizebox{\linewidth}{!}{
\begin{tabular}{|l|c|c|c|c|c|c|c|c|c|}
\hline
\textbf{95\% CI} & \textbf{DemosQA} & \textbf{GPCR} & \textbf{INCLUDE} & \begin{tabular}{@{}c@{}}\textbf{Greek ASEP} \\ \textbf{MCQA}\end{tabular}& \begin{tabular}{@{}c@{}}\textbf{Greek Medical} \\ \textbf{MCQA}\end{tabular} & \textbf{Plutus QA} & \begin{tabular}{@{}c@{}}\textbf{Greek} \\ \textbf{Truthful QA}\end{tabular} & \begin{tabular}{@{}c@{}}\textbf{Greek MMLU} \\ \textbf{(Greek-specific)}\end{tabular} & \textbf{CulturaQA}  \\
\hline
Krikri 8B & [-2.67\%, 5.33\%] & \textbf{[2.88\%, 16.83\%]} & \textbf{[3.44\%, 13.04\%]} & \textbf{[1.25\%, 7.08\%]} & [-1.39\%, 9.49\%] & \textbf{[2.22\%, 16.00\%]} & [-5.26\%, 2.33\%] & \textbf{[5.55\%, 8.66\%]} & \textbf{[0.22\%, 1.15\%]} \\
\hline
Plutus 8B & \textbf{[0.67\%, 9.50\%]} & \textbf{[7.69\%, 21.15\%]} & \textbf{[5.62\%, 15.04\%]} & \textbf{[1.42\%, 7.33\%]} & \textbf{[4.63\%, 15.74\%]} & \textbf{[9.33\%, 22.67\%]} & \textbf{[15.18\%, 22.64\%]} & \textbf{[6.20\%, 9.37\%]} & \textbf{[4.00\%, 5.10\%]}\\
\hline
EuroLLM v2 9B & \textbf{[4.17\%, 14.50\%]} & \textbf{[1.44\%, 19.71\%]} & \textbf{[14.31\%, 24.82\%]} & \textbf{[17.83\%, 24.50\%]} & \textbf{[12.04\%, 23.61\%]} & \textbf{[22.67\%, 38.67\%]} & \textbf{[12.73\%, 20.44\%]} & \textbf{[18.14\%, 21.86\%]} & \textbf{[1.21\%, 2.11\%]}\\
\hline
Gemma 3n E4B & [-0.67\%, 8.00\%] & [-3.85\%, 12.50\%] & \textbf{[3.99\%, 13.41\%]} & \textbf{[6.50\%, 12.50\%]} & \textbf{[0.69\%, 11.11\%]} & \textbf{[11.56\%, 27.11\%]} & \textbf{[3.06\%, 10.16\%]} & \textbf{[5.19\%, 8.39\%]} & \textbf{[2.44\%, 3.35\%]}\\
\hline
Qwen 3 8B & [-2.17\%, 6.17\%] & \textbf{[24.52\%, 40.87\%]} & \textbf{[4.71\%, 14.13\%]} & \textbf{[9.50\%, 15.75\%]} & \textbf{[7.87\%, 18.52\%]} & \textbf{[3.11\%, 16.44\%]} & \textbf{[7.22\%, 14.20\%]} & \textbf{[9.04\%, 12.16\%]} & \textbf{[2.82\%, 3.70\%]}\\
\hline 

\end{tabular}
}
\caption{\label{tab:7}Statistical significancy tests for Maistros 8B against the other open-weights LLMs. Results where the entire confidence interval is above zero are highlighted in bold.}
\end{table}

\section*{Discussion}

This study addresses the limited availability of resources for Greek QA and the performance gap of Greek-capable LLMs. We introduced \textit{CulturaQA}, a synthetic and human-curated dataset designed to support Greek LLM training and evaluation, and developed \textit{Maistros-8B}, a Greek-adapted open-weight LLM via knowledge distillation and fine-tuning. In addition, we proposed a memory-efficient evaluation framework and conducted a comprehensive assessment across multiple human-curated Greek QA datasets. The results provide evidence that curated synthetic data, combined with targeted fine-tuning, can improve the performance of open-weight models in under-resourced language settings.

\noindent Our empirical findings provide several insights aligned with the research questions. Specifically:
\begin{itemize}

\item The results indicate that high-quality QA datasets for training and evaluation can be constructed using synthetic data generated by LRMs and subsequently refined through human curation. The performance improvements observed for \textit{Maistros-8B} relative to its base model further support the utility of \textit{CulturaQA} (\textbf{RQ1} and \textbf{RQ2}).

\item Among the evaluated open-weight models, \textit{Maistros-8B} achieves consistently strong performance across most datasets, suggesting the effectiveness of the proposed knowledge distillation and fine-tuning approach (\textbf{RQ2} and \textbf{RQ3}).

\item However, open-weight models remain below the performance of proprietary ones for Greek QA. In particular, models such as \textit{Gemini 3 Flash} and \textit{GPT-5 Mini} consistently outperform all evaluated open-weight models across the considered benchmarks (\textbf{RQ4}).

\item The largest performance gain for \textit{Maistros-8B} is observed on \textit{PlutusQA}, a domain-specific dataset focused on the Greek economy, where it achieves an improvement of approximately 8\% over its base model and approaches the performance of \textit{GPT-5 Mini} (3.56\% difference). This suggests that the proposed approach is particularly effective in domain-specific settings.

\end{itemize}

This study has several limitations that suggest directions for future work. First, the analysis is restricted to Modern Greek, and the findings are not directly evaluated in other linguistic settings. Second, despite recent progress, there remains a limited number of Greek QA datasets, particularly those containing long-form answers and diverse content across both general and domain-specific topics. Third, our evaluation focuses on relatively small multilingual LLMs, as large-scale Greek-adapted models (i.e., exceeding 9B parameters) are currently scarce, which constrains broader comparisons. Finally, the proposed model is evaluated primarily on knowledge-intensive QA tasks. Other important capabilities, such as safety alignment and instruction following, are not explicitly assessed and remain an area for future investigation.

Overall, this work provides a foundation for future research in Greek QA and LLM adaptation for under-resourced languages. By releasing the dataset, model, and code, we aim to support the development of linguistically accurate and culturally grounded language models for Greek and related settings \cite{chang2025globalpiqaevaluatingphysical}. Future work may investigate the construction of more diverse post-training datasets to better capture the morphological and syntactic complexity of Modern Greek. This includes extending coverage to additional language variants, such as ancient and polytonic Greek \cite{kaddas_2023}, as well as regional dialects, which may further improve the modeling of social, historical, and cultural context \cite{chatzikyriakidis2026grddextendedgreekdialectal}. Finally, future work could extend the evaluation to domain-specific and long-context tasks, such as Greek legal QA \cite{chlapanis-etal-2025-greekbarbench, tsourma_2025}, as well as other NLP tasks including text summarization \cite{koniaris_2023, giarelis_2024_greekt5, giarelis_2024_greek_wiki} and text classification \cite{loukas-etal-2025-gr, liapis_2024, stylianou_2024, mastrokostas_2024}, aiming to provide a more comprehensive assessment of model capabilities and performance.

\section*{Ethical Considerations }
The source code, dataset, and generative model (research items) introduced in this work are intended solely for research and educational purposes. While the authors have made reasonable efforts to ensure the accuracy and reliability of the research items; these are provided without a warranty of any kind, regarding their suitability for any particular purpose. Moreover, the dataset introduced in this study is synthetically generated and was manually processed by the authors to remove any inappropriate or uninformative material, while trying to uphold ethical and high-quality data curation practices. 

\bibliography{references}

\section*{Acknowledgements}
This work has received funding from the European Union's Horizon Europe research and innovation programme under grant agreement No. 101235708 (BLUEPRINT – Building Living Urban Ecosystems through Participatory Renovation and Innovation Tools). Views and opinions expressed are however those of the author(s) only and do not necessarily reflect those of the European Union or the European Commission. Neither the European Union nor the European Commission can be held responsible for them.

\section*{Author contributions}

\textbf{N.G.:} Conceptualization, Data curation, Formal analysis, Investigation, Methodology, Resources, Software, Validation, Visualization, Writing – original draft, Writing – review \& editing. \\
\textbf{C.M.:} Conceptualization, Data curation, Investigation, Methodology, Software, Writing – review \& editing. \\
\textbf{N.K.:} Project administration, Funding acquisition, Supervision, Writing – review \& editing. \\All authors reviewed the manuscript. 

\section*{Data availability}
The proposed model and dataset are available in \url{https://huggingface.co/IMISLab}. The other evaluation datasets used in the study can be found in \url{https://huggingface.co/datasets}.

\section*{Code availability}
The source code of this study is available in \url{https://github.com/NC0DER/Maistros}.

\section*{Competing Interests Statement}
The authors declare no competing interests.

\setcounter{table}{0} 
\renewcommand{\thetable}{A\arabic{table}} 
\section*{Appendix A} 

\renewcommand{\arraystretch}{1.3} 
\begin{longtable}{|p{0.15\linewidth}|p{0.38\linewidth}|p{0.38\linewidth}|}
\caption{\label{tab:prompts}The prompts utilized in this study alongside their English translations.} \\
\hline
\textbf{Prompt Type} & \textbf{Greek} & \textbf{English (Translated)} \\
\hline
\endfirsthead

\multicolumn{3}{c}{{\tablename\ \thetable{} -- continued from previous page}} \\
\hline
\textbf{Prompt Type} & \textbf{Greek} & \textbf{English (Translated)} \\
\hline
\endhead

\hline \multicolumn{3}{|r|}{{\textit{Continued on next page}}} \\ \hline
\endfoot

\hline
\endlastfoot

Dataset Creation (Questions) & 
\textgreek{Είσαι ένα εξαιρετικά ανεπτυγμένο μοντέλο Τεχνητής Νοημοσύνης για την Ελληνική γλώσσα.}\newline
\textgreek{Χρησιμοποίησε τις παρακάτω οδηγίες για να δημιουργήσεις μια σειρά ερωτήσεων στο θέμα που αναφέρει ο χρήστης:}\newline
\newline
\textgreek{Οδηγίες:}\newline
1. \textgreek{Απάντα αποκλειστικά στα Ελληνικά με άψογη γραμματική, σύνταξη και ορθογραφία.}\newline
2. \textgreek{Λάβε υπόψη τον ελληνικό πολιτισμό και την ελληνική κοινωνική πραγματικότητα όπου είναι σχετικό.}\newline
3. \textgreek{Απόφυγε τη χρήση στερεοτύπων.}\newline
4. \textgreek{Βάλε πάντα το σύμβολο} \textbullet{} \textgreek{πριν από κάθε ερώτηση.}\newline
5. \textgreek{Δημιούργησε σημαντικές, συχνά προκύπτουσες και χρήσιμες ερωτήσεις για το θέμα.}\newline
6. \textgreek{Όλες οι ερωτήσεις πρέπει να μπορούν να απαντηθούν αντικειμενικά.}\newline
7. \textgreek{Μη δημιουργείς επαναλαμβανόμενες ερωτήσεις.}\newline
8. \textgreek{Κάθε ερώτηση πρέπει να είναι σαφώς ορισμένη.}\newline
9. \textgreek{Γράψε μόνο το κείμενο των ερωτήσεων, χωρίς επιπλέον σχόλια.}\newline
\newline
\textgreek{Παρακαλώ δημιούργησε 15 ερώτησεις για το εξής θέμα:} \{topic\} & 
You are an extremely developed Artificial Intelligence model for the Greek Language.\newline
Use the following instructions to create a series of questions on the topic mentioned by the user:\newline
\newline
Instructions:\newline
1. Answer exclusively in Greek with impeccable grammar, syntax and spelling.\newline
2. Take into consideration the Greek civilization and the Greek social reality where relevant.\newline
3. Avoid the use of stereotypes.\newline
4. Always place the symbol \textbullet{} before each question.\newline
5. Create significant, frequently occurring and useful questions for the topic.\newline
6. All questions must be able to be answered objectively.\newline
7. Do not create repeated questions.\newline
8. Every question must be clearly defined.\newline
9. Write only the text of the questions, without extra comments.\newline
\newline
Please create 15 questions for the following topic: \{topic\} \\
\hline

Dataset Creation (Answers) & 
\textgreek{Είσαι ένα εξαιρετικά ανεπτυγμένο μοντέλο Τεχνητής Νοημοσύνης για την Ελληνική γλώσσα.}\newline
\textgreek{Χρησιμοποίησε τις παρακάτω οδηγίες για να παράγεις τη καλύτερη δυνατή απάντηση:}\newline
\newline
\textgreek{Οδηγίες:}\newline
1. \textgreek{Απάντα αποκλειστικά στα Ελληνικά με άψογη γραμματική, σύνταξη και ορθογραφία.}\newline
2. \textgreek{Λάβε υπόψη τον ελληνικό πολιτισμό και την ελληνική κοινωνική πραγματικότητα όπου είναι σχετικό.}\newline
3. \textgreek{Απάντησε στις ερωτήσεις του χρήστη με ειλικρίνεια και επιστημονική ακρίβεια.}\newline
4. \textgreek{Απόφυγε τη χρήση στερεότυπων.}\newline
5. \textgreek{Αν η ερώτηση είναι ασαφής ή λείπει πληροφορία (π.χ. χώρα, περίοδος):}\newline
-- \textgreek{Μην ζητήσεις διευκρίνιση.}\newline
-- \textgreek{Δώσε την απάντηση κάνοντας ρητές παραδοχές (π.χ. "Ελλείψει άλλης αναφοράς, θεωρούμε ως προεπιλογή την Ελλάδα και το τρέχον έτος").}\newline
\newline
\textgreek{Παρακαλώ απάντησε στη παρακάτω ερώτηση:} \{question\} & 
You are an extremely developed Artificial Intelligence model for the Greek Language.\newline
Use the following instructions to generate the best possible answer:\newline
\newline
Instructions:\newline
1. Answer exclusively in Greek with impeccable grammar, syntax and spelling.\newline
2. Take into consideration the Greek civilization and the Greek social reality where relevant.\newline
3. Answer the user question with honesty and scientific accuracy.\newline
4. If the question is vague or information is missing (e.g., country, time period):\newline
-- Do not ask for clarification.\newline
-- Give the answer by making explicit assumptions (e.g. "In the absence of other reference, we assume as a default Greece and the current year”).\newline
\newline
Please answer the following question: \{question\} \\
\hline

Multiple Choice\newline (Evaluation) & 
\textgreek{Διάβασε προσεκτικά την ερώτηση και σκέψου ποια επιλογή είναι σωστή.}\newline
\textgreek{Επίλεξε την καλύτερη απάντηση.}\newline
\textgreek{Απάντησε μόνο με το γράμμα (Α, Β, Γ ή Δ).}\newline
\newline
\textgreek{Ερώτηση:} \{question\}\newline
\newline
\textgreek{Απαντήσεις:}\newline
\{answers\} & 
Read the question carefully and think about which option is correct.\newline
Choose the best answer.\newline
Answer only with the letter (A, B, C or D).\newline
\newline
Question: \{question\}\newline
\newline
Answers:\newline
\{answers\} \\
\hline

Open-ended\newline (Evaluation) & 
\textgreek{Σκέψου και απάντησε στην παρακάτω ερώτηση με συντομία, σαφήνεια και ακρίβεια.}\newline
\newline
\textgreek{Ερώτηση:} \{question\} & 
Think and answer the following question with brevity, relevance and precision.\newline
\newline
Question: \{question\} \\
\end{longtable}
\end{document}